\documentclass[runningheads]{llncs}

\usepackage[T1]{fontenc}
\usepackage[utf8]{inputenc}
\usepackage{lmodern}
\usepackage{color}
\usepackage{hyperref}
\usepackage{microtype}
\usepackage{graphicx}
\usepackage{amsmath}
\usepackage{tabularx}
\usepackage{array}
\usepackage{longtable}

\microtypesetup{expansion=false}
\newcolumntype{L}[1]{>{\raggedright\arraybackslash}p{#1}}
\newcolumntype{Y}{>{\raggedright\arraybackslash}X}

\hypersetup{
  colorlinks=true,
  linkcolor=black,
  citecolor=black,
  urlcolor=blue,
  pdftitle={Transparent Screening for LLM Inference and Training Impacts},
  pdfauthor={Arnault Pachot and Thierry Petit}
}

\title{Transparent Screening for LLM Inference and Training Impacts}
\titlerunning{Transparent Screening for LLM Impacts}
\author{Arnault Pachot\inst{1}\thanks{Corresponding author: Arnault Pachot, \email{apachot@emotia.com}} \and Thierry Petit\inst{1}}
\authorrunning{A. Pachot and T. Petit}
\institute{Emotia, Paris, France\\
\email{\{apachot,tpetit\}@emotia.com}}

\begin{document}

\maketitle

\begin{abstract}
This paper presents a transparent screening framework for estimating inference and training impacts of current large language models under limited observability. The framework converts natural-language application descriptions into bounded environmental estimates and supports a comparative online observatory of current market models. Rather than claiming direct measurement for opaque proprietary services, it provides an auditable, source-linked proxy methodology designed to improve comparability, transparency, and reproducibility.

\end{abstract}

\section{Introduction}
This paper documents the transparent screening framework used in the \emph{ImpactLLM Observatory}. The observatory now covers 41 models, and this manuscript explains how the interface extracts scenarios from natural-language descriptions, propagates them through bounded multi-factor screening proxies, and publishes the resulting comparative tables and figures. The goal is to keep those proxies auditable while keeping the application workflow fast and interpretable.

The main contributions of this paper are threefold. First, it introduces a transparent screening framework for estimating inference and training impacts of large language models under limited observability. Second, it proposes a bounded multi-factor proxy methodology that separates inference and training estimates while making the main assumptions explicit. Third, it provides an operational implementation and a comparative observatory covering current market models.

The remainder of this paper covers the technical design of the inference estimator, the training proxy, the comparative outputs currently exposed in the observatory, and the limits of those assumptions.

The methodological premise is straightforward. For the dominant hosted LLM services, direct provider-side environmental telemetry is generally unavailable, even though those systems dominate practical use and public debate. In that context, avoiding approximation altogether does not produce a more rigorous discussion; it leaves room for opaque claims, unsupported comparisons, and contradictory numerical narratives. The role of the present paper is therefore not to claim direct measurement where none exists, but to define a bounded, source-linked, inspectable proxy framework that is methodologically preferable to unverifiable assertions.

\section{Inference Screening Method}
The main practical question addressed by the tool is straightforward: given a software feature using an LLM, can we produce an inference estimate that is useful for screening and comparison when direct provider telemetry is absent?

The current answer is deliberately limited. The estimator is \emph{inference-only}. It excludes model training, embodied impacts, application-side software consumption, and ancillary infrastructure from the displayed result. This is not because those dimensions are unimportant, but because the available inference anchors are structured enough to support a transparent screening method, whereas mixing them with broader lifecycle terms would blur the meaning of the result.

The design objective is also practical speed. Instead of asking users to fill a large technical form, the web interface starts from a natural-language description such as ``We use GPT-4o-mini for customer support, around 4{,}000 uses per month.'' A parser then maps this text to a compact scenario with model, request type, approximate tokens, usage volume, and country assumptions. The output is therefore fast to obtain, but every inferred parameter remains visible to the user and can be inspected or challenged.

The current market-model release follows five rules.
\begin{itemize}
\item It starts from an observed literature anchor rather than from provider claims. The retained prompt-energy anchor is the Gemini Apps median prompt energy reported by Elsworth et al.: $0.24$ Wh/prompt \cite{elsworth2025}.
\item It makes token volume explicit. Prompt compute is approximated from input and output tokens, with a larger weight for output generation than for input processing.
\item It does not scale on raw parameters alone. It constructs \emph{effective active parameters} by adjusting the target profile with context-window class, serving mode, modality support, and architecture notes.
\item It reports a bounded low-central-high interval rather than a single falsely precise point.
\item It derives carbon from retained energy and the electricity mix of the retained country, which is a contextual assumption rather than a model measurement.
\end{itemize}

Formally, let $E_a = 0.24$ Wh be the anchor value and $P_a = 180$ the active-parameter proxy retained for the anchor model. For a target model $t$, let $P_t$ denote active parameters, $T_{in}$ input tokens, and $T_{out}$ output tokens. We define weighted prompt-compute volume:
\begin{equation}
V_t = T_{in} + \omega T_{out},
\end{equation}
with $\omega = 1.8$ in the current release. The LLM inference literature distinguishes a prompt prefilling stage from an autoregressive decoding stage for output generation \cite{zhou2024survey}. Here, $\omega = 1.8$ is used as an internal screening prior to reflect that asymmetry, not as a directly measured constant from the cited source. The project reference scenario is:
\begin{equation}
V_{ref} = 1000 + 1.8 \times 550 = 1990.
\end{equation}

Effective active parameters are then defined by:
\begin{equation}
P^{eff}_{t,s} = P_t \times F^{ctx}_{t,s} \times F^{srv}_{t,s} \times F^{mod}_{t,s} \times F^{arch}_{t,s},
\end{equation}
where $s \in \{\mathrm{low}, \mathrm{central}, \mathrm{high}\}$ indexes the screening scenario. The final per-request energy estimate is:
\begin{equation}
\hat{E}_{t,s} = E_a \times \left(\frac{P^{eff}_{t,s}}{P_a}\right)^{\alpha_s}
\times \left(\frac{V_t}{V_{ref}}\right)^{\beta_s}.
\end{equation}
The current implementation uses $\alpha_{low}=0.85$, $\alpha_{central}=0.95$, $\alpha_{high}=1.05$, and $\beta_{low}=0.85$, $\beta_{central}=0.92$, $\beta_{high}=1.0$. These exponents are internal screening priors. The central values are set close to linearity, but slightly below 1, to preserve interpretability while avoiding overstatement of cross-model differences from a single transferred anchor. The low and high values widen the response to reflect uncertainty in how effective active parameters and weighted token volume map to inference energy across opaque production systems. Carbon is then derived from the retained country mix:
\begin{equation}
\hat{C}_{t,s,c} = \frac{\hat{E}_{t,s}}{1000} \times CI_c,
\end{equation}
where $CI_c$ is the carbon intensity of country $c$ in gCO$_2$e/kWh.

This method is not presented as a physical law of inference scaling. It is a traceable screening proxy anchored in observed prompt energy and explicit contextual assumptions. Its purpose is to provide a fast and inspectable estimate for comparative reasoning, not an audited declaration of real provider-side energy use.

\section{Training Screening Method}
The observatory also exposes training orders of magnitude for current market models. Here the uncertainty is even higher than for inference because published values are sparse, heterogeneous, and often reported only as aggregate emissions. A pure parameter-only scaling is therefore too brittle. The current release instead uses a second bounded proxy that combines retained parameter count with additional training priors.

The training proxy starts from literature anchors that directly report training CO$_2$e, and from training-energy reconstructions derived from those emissions when the source-country electricity mix is documented. For a target model $t$, the central training estimate is:
\begin{equation}
\hat{E}^{train}_{t,s} = E^{train}_a \times \left(\frac{P_t}{P_a}\right)^{\alpha_s}
\times \left(\frac{Tok_t}{Tok_a}\right)^{\beta_s}
\times F^{reg}_{t,s} \times F^{arch-tr}_{t,s} \times F^{hw}_{t,s},
\end{equation}
where $P_t$ is the retained parameter count, $Tok_t$ is a training-token prior, $F^{reg}$ is a training-regime prior, $F^{arch-tr}$ captures architecture and multimodality assumptions, and $F^{hw}$ is a hardware-class proxy. The value of 20 training tokens per retained parameter is currently used as a project screening prior when no model-specific public estimate is available. It should be read as a pragmatic default assumption, not as a direct measurement. This default is broadly consistent with the order of magnitude illustrated by Chinchilla, whose compute-optimal example combines 70B parameters with 1.4T training tokens, i.e. about 20 tokens per parameter \cite{hoffmann2022chinchilla}. The current release also defaults market models to a foundation-pretraining regime unless a narrower public indication exists.

This second proxy is not used in the application estimator shown to users, which remains inference-only. It is used in the observatory and the comparative tables to avoid suggesting that training impacts can be projected from parameter count alone. The result should still be read as a screening order of magnitude rather than as an audited declaration of model-development impact.

\section{Results and Comparative Outputs}
The reported values reflect the recalculated inference and training columns for the expanded market-model catalog. The observatory emphasizes the central prompt-level proxy per standardized request before any application-specific annualization.

Table \ref{tab:short-results-method} summarizes a few illustrative outputs. The first block shows how the standardized observatory makes model orders of magnitude legible. The second block shows how the same proxy can be annualized at the software-feature level from compact, natural-language scenarios.

\begin{table}[htbp]
\centering
\small
\begin{tabularx}{\textwidth}{|L{4.0cm}|L{2.5cm}|L{2.2cm}|Y|}
\hline
Case & Energy & Carbon & Interpretation \\
\hline
Ministral 3B, one standardized request & 0.0056 Wh & 0.0002 gCO$_2$e & Small hybrid model under a French provider-country proxy \\
\hline
GPT-5 mini, one standardized request & 0.1706 Wh & 0.0657 gCO$_2$e & Medium proprietary hosted model under a US provider-country proxy \\
\hline
Claude Opus 4.1, one standardized request & 2.9922 Wh & 1.1520 gCO$_2$e & Very large proprietary estimate illustrating the steep growth of screening orders of magnitude \\
\hline
Support chatbot, Ministral 8B, 20{,}000 conversations/month & 2.38 kWh/year & 96 gCO$_2$e/year & Low-carbon annual result despite high usage because the retained electricity factor is favorable \\
\hline
Retrieval assistant, GPT-5 mini, 4{,}000 uses/month & 12.31 kWh/year & 4.74 kgCO$_2$e/year & Higher annualized result driven by hosted-service assumptions and US electricity contextualization \\
\hline
\end{tabularx}
\caption{Illustrative outputs from the current observatory and application estimator. These values are screening estimates, not audited declarations.}
\label{tab:short-results-method}
\end{table}

These examples illustrate three practical properties of the method. First, annualization matters: apparently small unit values become relevant when multiplied by real usage. Second, the chosen model profile matters materially. Third, the retained electricity mix can dominate the carbon interpretation, which is why country contextualization is explicitly shown in the interface. Just as importantly, these results can be obtained quickly from natural-language inputs while keeping the extracted scenario and assumptions visible in the interface.

To keep the presentation focused, Table \ref{tab:representative-models} reports a reduced set of representative models spanning the low, medium, and high ends of the current catalog. The full catalog is omitted here for space reasons and can be provided as supplementary material or released with the public observatory.

\begin{table}[htbp]
\centering
\small
\begin{tabularx}{\textwidth}{|L{3.5cm}|L{1.8cm}|L{2.2cm}|L{1.4cm}|Y|}
\hline
Model & Inference Wh/request & Inference gCO$_2$e/request & Training GWh & Interpretation \\
\hline
Claude Opus 4.1 & 2.9922 & 1.1520 & 125.63 & Upper-end proprietary model with very large screening values for both inference and training \\
\hline
GPT-5.2 & 2.7897 & 1.0740 & 101.76 & Frontier proprietary model illustrating the high end of current screening estimates \\
\hline
Gemini 2.5 Pro & 0.3601 & 0.1387 & 1.26 & Upper-mid-range model with markedly lower screening values than the top tier \\
\hline
GPT-5 mini & 0.1706 & 0.0657 & 0.28 & Medium model representative of lighter hosted deployments \\
\hline
Llama 3.1 70B & 0.1043 & 0.0402 & 0.16 & Large open-weight model with moderate inference estimates \\
\hline
GPT-4o mini & 0.0155 & 0.0060 & 0.0027 & Lightweight hosted model with low per-request screening impacts \\
\hline
Ministral 3B & 0.0056 & 0.0002 & 0.0004 & Low-end model illustrating the lower bound of the current catalog \\
\hline
\end{tabularx}
\caption{Representative subset of observatory results spanning the current market-model catalog. Inference values correspond to the central prompt-level proxy retained by the observatory, expressed per standardized request. Training values are multi-factor screening estimates; additional decimals are shown for small training values to avoid false zeros from rounding.}
\label{tab:representative-models}
\end{table}

\section{Related Work}
The results above are not produced from a single benchmark, but from a layered source base combining direct environmental anchors, estimator literature, infrastructure and electricity context, and broader analytical framing. This structure corresponds to the source logic exposed in the site's \emph{Sources} page, where each value is linked to a document, a system boundary, and a use role in the estimator.

\par Unlike prior work focused on direct telemetry, training case studies, or infrastructure-side accounting tools, the proposed framework targets comparative screening under partial observability for currently deployed LLM services.

\subsection{Direct Environmental Anchors for LLMs}
The primary literature anchors for training and inference come from work that reports environmental indicators directly for language-model systems or adjacent production systems. The core training anchors include Strubell et al. on transformer training costs \cite{strubell2019}, Luccioni et al. on BLOOM \cite{luccioni2023}, Morrison et al. on holistic language-model impacts \cite{morrison2025}, and recent lifecycle or training-oriented extensions such as Fernandez et al. \cite{fernandez2025lca} and d'Orgeval et al. \cite{dorgeval2026}. For inference, the most operational prompt-level anchor in the current release is Elsworth et al. \cite{elsworth2025}, complemented by broader framing from Ren et al. \cite{ren2024} and water-related contextualization from Li et al. \cite{li2025water}.

\subsection{Measurement, Tracking, and Proxy Literature}
The methodological design also builds on the literature and tooling ecosystem for measuring or estimating AI impacts when direct telemetry is available or partially missing. This includes CarbonTracker \cite{anthony2020carbontracker}, CodeCarbon \cite{codecarbon2026}, the ML.ENERGY benchmark \cite{mlenergy2025}, and proxy-oriented tooling such as EcoLogits \cite{ecologits2026}. These references do not provide one transferable number for all current hosted LLMs, but they define the design space within which a screening proxy must remain explicit about assumptions, units, and uncertainty.

\subsection{Infrastructure, Electricity, and Cloud Context}
A second family of sources informs the infrastructural and contextual layers needed to convert energy proxies into carbon estimates or to interpret data-center scale effects. This includes cloud and infrastructure accounting references such as Cloud Carbon Footprint \cite{cloudcarbonfootprint2026}, AWS CCFT \cite{awsccft2026}, the Microsoft Emissions Impact Dashboard \cite{azureeid2026}, Google Cloud carbon reporting \cite{gcloudcarbon2026}, and the OVHcloud environmental tracker \cite{ovhtracker2024}. It also includes system-level or electricity-demand framing from EPRI \cite{epri2024}, Lawrence Berkeley National Laboratory \cite{lbl2025,shehabi2024report}, the International Energy Agency \cite{iea2025}, and Joule-scale discussion by de Vries-Gao \cite{devriesgao2025joule}.

\subsection{Model Documentation and Conceptual Framing}
Some source families play a different role: they do not provide direct environmental measurements but document model properties or frame the interpretation of the results. Examples include model and provider documentation such as the Llama 3.1 model card \cite{meta2024llama}, broader environmental framing from Rillig et al. \cite{rillig2023}, and earlier conceptual work on sustainable AI and environmental decision support \cite{pachot2022book,pachot2023sustainableai}. The observatory and the site itself are part of this transparency layer because the project republishes those linked records in reusable interfaces \cite{impactllmrepo2026}.

\section{Discussion and Perspectives}
Beyond the catalog table and the static release timelines, the observatory can also be read as a dynamic screening signal about the acceleration of frontier-model impacts. Even at inference time, model families can diverge rapidly in their comparative screening values as capability tiers rise, while training estimates still dominate the highest orders of magnitude.

This reading remains interpretive rather than declarative. The observatory does not only contain isolated point estimates; it also suggests a market trajectory in which the environmental stakes of frontier models may be rising rapidly. But the value of that reading depends entirely on the transparency of the underlying assumptions. If those assumptions change, the comparative gaps change as well. The observatory should therefore be read as a discussion tool and policy prompt, not as a direct provider-side measurement or an immutable forecasting law.

From a policy and governance perspective, this is arguably one of the most useful roles an open observatory can play. Even without direct industrial telemetry, it can surface the possibility that model-scale growth is outpacing the public visibility of environmental reporting. In that sense, the discussion value of the observatory is not limited to ranking models; it also lies in making structural trends legible enough to motivate better disclosure, more granular reporting, and more disciplined debate.

\section{Limitations}
The main limitations are methodological. First, the estimator depends on scarce literature anchors, so the uncertainty intervals rely on wide parameter- and token-exponent bounds to keep the screening range honest. Second, proprietary service configurations remain opaque, forcing the method to rely on contextual proxies for regime, architecture, and hardware. Third, training-level emissions are reconstructed from sparse reports and assumed token priors; more published training anchors would reduce that uncertainty. Finally, the natural-language parser must interpret ambiguous descriptions, which is why every extracted assumption is surfaced to the user along with the final estimate.

\section{Implementation Notes}
The prototype is implemented as a Python/Flask web application that builds on a structured observatory dataset. The front-end sends the user description to a language-model parser, then queries the estimator to compute the inference and training proxies. The same dataset drives the observatory tables, which remain independent of the estimator logic.

\section{Conclusion}
The contribution of this manuscript is both methodological and documentary: it explains how the system transforms natural-language descriptions into bounded inference and training estimates, and it makes the resulting comparative tables and figures for the tracked models directly inspectable in one place.

That contribution should not be read as an attempt to eliminate uncertainty. For the most widely used proprietary LLM services, uncertainty is structurally unavoidable because the decisive telemetry remains inaccessible. The relevant methodological task is therefore to organize approximation rather than to pretend to escape it. This paper argues that a transparent, source-linked, reproducible screening proxy is a better basis for comparison and discussion than the current situation in which strong environmental claims often circulate without explicit assumptions, traceable derivation, or shared comparison rules.

\bibliographystyle{splncs04}
\bibliography{references_ImpactLLM}
\end{document}